# Shape-based defect classification for Non Destructive Testing

Gianni D'Angelo, Salvatore Rampone
University of Sannio
Dept. of Science and Technology
Benevento, Italy
{dangelo, rampone}@unisannio.it

*Abstract*— The aim of this work is to classify the aerospace structure defects detected by eddy current non-destructive testing. The proposed method is based on the assumption that the defect is bound to the reaction of the probe coil impedance during the test. Impedance plane analysis is used to extract a feature vector from the shape of the coil impedance in the complex plane, through the use of some geometric parameters. Shape recognition is tested with three different machine-learning based classifiers: decision trees, neural networks and Naive Bayes. The performance of the proposed detection system are measured in terms of accuracy, sensitivity, specificity, precision and Matthews correlation coefficient. Several experiments are performed on dataset of eddy current signal samples for aircraft structures. The obtained results demonstrate the usefulness of our approach and the competiveness against existing descriptors.

## I. INTRODUCTION

In aircraft manufacturing and maintenance, Non-Destructive Testing (NDT) is widely used for evaluating the property of materials, components, systems, without causing damage during the analysis. In aerospace industry, NDT is not only necessary but plays a crucial role in the risk management.

There are many NDT technologies that are available for aircraft inspections. The most common methods are based on visual and optical testing: optical holography, X-ray, ultrasonic wave, infrared detection, X-ray and ultrasonic C-scan.

Eddy Current Testing (ECT) [1] is one of the NDT methods that are widely used in detecting surface or subsurface crack of conductive materials. Recent technologies have made ECT more powerful and useful in quality assurance. Modern ECT techniques are based on low-cost methods and are easily applicable for the inspection where premature failures could give rise to economic or related to human life problems. However, most of the ECTs in maintenance, are currently conducted by human inspectors, whose individual experiences may yield to differences in the interpretation of the results. Generally, the human operators take decision on the acceptability of the piece under test by evaluating typical bidimensional graphs that depict the reactance versus the resistance of the coil impedance of the ECT probe. Reliable human performance is vital to inspections and tests. Inadequate human performance could lead to missed defects and inaccurate reports, with potentially serious safety and cost consequences. In addition, manpower assigned to such tasks imply significant recurrent costs and it is time consuming in itself. The presence of defects and damages in aerospace structure pose a significant threat to the integrity of the structures and also in the reliability and safety of the inspections. For this reason, the accuracy of diagnosis of aerospace materials cannot be left only to the human capacity but also entrusted to mathematical models and advanced methods of data processing [2]. Different solutions have been proposed to address this issue [3-5]. The main methods are based on two principal approaches: the *model-based* and the *model-free* techniques. The first refers to the use of computer simulators to determine the material electromagnetic parameters by measuring the probe impedance and relying on mathematical models of probe geometry [6]. The second are based on the use of signal analysis acquired during the test. In this approach, the methods of spectrum analysis and pattern recognition are often used in multi-parametric control for data processing [7]. The signals acquired from the sensors are processed in the frequency domain to extract the magnetic field information. In fact, the presence of a defect causes an asymmetricity in the induced signals that can be highlighted in the frequency domain by analyzing the signal harmonics [8]. In [9] the authors proposed a "model-free" method based on the spectrum analysis and on a proper algorithm that uses a machine learning technique [10]. However, the application of these methods requires sophisticated techniques for processing signals that, for a large amount of date, lead to have long process time [11]. These do not allow the automation of the test and deprive them of the same dynamism typical of a system able to adapt itself to changes in the parameters of the testing system at run-time. NDT should be performed with methods able to collect the most comprehensive information about new defects, expand existed base of defects and increase diagnostics system precision in runtime. The key to a successful testing system is to choose the right features extraction method that represent the defect as accurately and uniquely as possible in a short time. Recently, the image analysis and pattern recognition techniques in NDT are being increasingly used to increase the objectivity, consistency and efficiency of the testing [12, 13]. This new "model-free" based approach involves extracting of the hidden useful "knowledge" [14-16] embedded in the images representing the ECT results. The fast growth of multimedia data [17], mainly due to the wide spread of digital devices [18], has lead to the developing of high-performance techniques in the field of image retrieval. These techniques are generally known as Content-Based Image Retrieval (CBIR) systems [19]. CBIR systems perform the image retrieval through a similarity process which is defined in terms of visual features with more objectiveness. The Shape Geometric Description (SGD) [20] of the objects, represented in images, is one of the most significant properties used in CBIR tasks. One of the main advantages of this approach is the performance in terms of velocity. In fact, the feature data for each of the visual attributes of each image is very much smaller in size compared to the image data.

In this paper we propose a CBIR system to recognize the image that assumes the impedance of the EC probe coil in the complex plane for a specific defect. We make use of some geometric parameters to descript the shape of the probe coil impedance. Shape recognition is tested on the automated classification of eddy current signatures with three different machine-learning based classifiers: decision trees, neural networks and Naive Bayes. The performance of the proposed detection system are measured in terms of accuracy, sensitivity, specificity, precision and Matthews correlation coefficient.

The paper is organized as follows. In Section 2 we briefly give a description of the image retrieval issue. Discussions related to the features extraction from the EC impedance shape in the complex plane are presented in section 3. In Section 4 we explain the SGD-based proposed experiments and we show the related results. Last section is devoted to the conclusions.

## II. IMAGE RETRIEVAL

### A. CBIR

Content Based Image Retrieval (CBIR) is an actively researched area in computer vision whose goal is to find images similar in visual content to a given query from an image dataset [21]. Image analysis can be based on several distinct features such as color [22], texture [23], shape [24] or any other information that can better describe the image. A typical CBIR system extracts the features from each image in the dataset and stores them in the database. Then, when similar images are searched using a "query" image (the image is used as a query), a feature vector is first extracted from this query image, then a distance between the calculated vector and the database image features is computed. Typical distance metrics between the feature vectors include: Canberra distance, Euclidean distance, Manhattan metric, Minkowski metric and others [25]. If the calculated distance is small, then the images compared are considered similar. Compared with the traditional methods, which represent image contents by keywords, the CBIR systems are faster and more efficient. The main advantage of CBIR system is that it uses image features rather than image itself. For these reason the application areas are numerous and different: remote sensing, geographic information systems, weather forecasting, medical imaging [26] and in the last few years also in image search on the Internet [27, 28].

There are many different implementations of CBIR, so that the CBIR has attracted the attention of researchers across several disciplines. Nevertheless, the key to a good retrieval system is to choose the right features that better represent the images while minimizing the computation complexity.

### B. Shape geometric descriptor (SGD)

The shape descriptor aims to measure geometric attributes of an image. There are many different kinds of shape matching methods, and the progress in improving the matching rate has been substantial in recent years. However, these descriptors are categorized into two main groups: *region-based* shape descriptors and *contour-based* shape descriptors [29]. The first method uses all the pixel information within a shape region of a image. Common region-based methods make use of moment descriptors [30] that include: geometric moments, Legendre moments, Zernike moments and others [31]. Contour-based approaches use only the information related to the boundary of a shape region and do not consider the shape interior content. These include Fourier descriptor, Wavelet descriptors, curvature scale space and shape signatures [32].

In Fig. 1 is depicted typical geometric parameters for the shape signatures. They include: Area (A), perimeter (P), centroid (G), orientation angle (α), principal inertia axes, width (W), length (L) and surfaces of symmetry ($S_i$) for an equivalent ellipse image region.

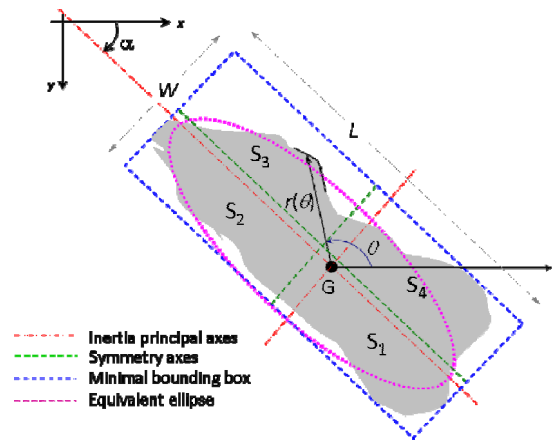

Fig. 1. Typical geometric parameters.

From these base parameters some advanced parameters (not changing when the original object is submitted to translation, scale changes and rotations) can be derived. They include [33]:

*Compactness*: $C=4\pi A/P^2$. It represents the ratio of the shape area to the area of a circle having the same perimeter.

*Elongation*: E=L/W. It is defined by the ratio of the length to the width of the minimal rectangle surrounding the object called also the minimal bounding box.

*Rectangularity*: R=A/(L x V). It represents how rectangular a shape is. It is equal to the ratio of the shape area to the area of its minimal bounding box.

*Eccentricity*: It represents the measure of aspect ratio. It is obtained from the ratio of the minor axis to the major axis in the object equivalent ellipse.

*Convexity*: It is defined as the ratio of perimeters of the convex hull over that of the original contour.

### III. PROPOSED FEATURE EXTRACTION METHOD

We investigate the application of the CBIR techniques to characterize the image that assumes the impedance of the EC probe coil in the complex plane for a specific defect.

Eddy current testing is based on the physics phenomenon of electromagnetic induction [34]. In an eddy current probe, an alternating current flows through a wire coil and generates an oscillating magnetic field. If the probe and its magnetic field are taken close to a conductive material like a metal test piece, a circular flow of electrons known as an eddy current will begin to move through the metal like swirling water in a stream. That eddy current flowing through the metal will in turn generate its own magnetic field, which will interact with the coil and its field through mutual inductance. Changes in metal thickness or defects like near-surface cracking will interrupt or alter the amplitude and pattern of the eddy current and the resulting magnetic field. This in turn affects the movement of electrons in the coil by varying the electrical impedance of the coil. In this approach, the presence of damage is characterized by the changes in the signature of the resultant signal that propagates through the structure. So, changing material parameters corresponds to a particular output impedance that is characterized by a specific shape in the complex plane. We use the shape of the impedance in the complex plane as signatures to identify changes in the test piece. In Fig. 2 a typical shape of coil impedance in the complex plane for an aluminum sample with notch perpendicular of width 0.3 mm and depth 1 mm is shown.

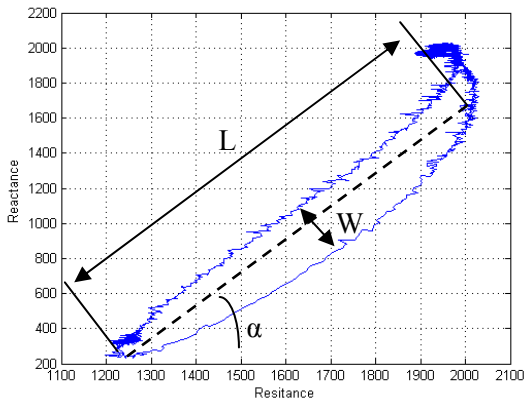

Fig. 2. Typical shape of coil impedance in the complex plane. The dotted line is principal inertia axis. (L,W,α) is the feature vector.

As depicted, we intercept the principal inertia axis and we use the set (L,W,α), composed by the length (L), width (W) and orientation angle (α) of the shape, as feature vector that characterizes a defect. These features are used as input to machine learning based classifiers.

### IV. EXPERIMENTS AND RESULTS

In order to evaluate the suitability and usefulness of the proposed set of features to describe shape of the coil impedance in the complex plane, we selected twelve types of defects, with 20 shape for each category (class). The classification performance was evaluated using a ten-fold cross-validation method.

#### A. Sample data

The data used in the study refers to a subset of a database with EC signal samples for aircraft structures [35]. The overall database is divided in 4 parts. The first contains 240 records acquired on an aluminum sample with notches of width 0.3 mm, depth 0.4, 0.7, 1, and 1.5 mm perpendicular, depth 0.4, 0.7, 1, and 1.5 mm with an angle of 30 degrees, 0.7, 1 and 1.5 mm with an angle of 60 degrees and 1.5 mm with an angle of 45 degrees. The second refers to 150 records, notches of width 0.2 mm, depth 1, 3 and 5 mm, both perpendicular and 45 degrees orientation of a stainless steel structure. Third database refers to two-layer aluminum aircraft structure with rivets, two notches below the rivets in the first layer (width 0.2 mm, length 2.5 mm, angle 90 degrees and 30 degrees) and two in the second layer (width 0.2 mm, length 2.5 mm and 5 mm, angle 90 degrees), two defect-free rivets. The fourth sample refers to four-layer aluminum structure (layer thickness 2,5 mm) with rivets containing 4 notches (width 0.2 mm, length 2.5 mm, angle 90 deg ) below the rivets in the first, second, third or fourth layer, four defect-free rivets.

In this work we used the first part of the database that contains 240 records acquired on an aluminum sample with notches which differ in the depth and angle. Each record contains 4096 samples, an sampling frequency of 10KHz and two canals (real and imaginary part) for each acquired measure.

#### B. Pre-Processing

To obtain an image suitable for feature extraction, we performed a noise reduction by removing the irrelevant information from the shapes. This was accomplished by detecting and extracting the image regions of interest, cropping them through their bounding box. As depicted in Fig. 2 the most noise is concentrated on the top-right side which represents a high value for both real and imaginary part of the coil impedance. This is confirmed by also the spectrum analysis of each single channel of the samples [9]. Using Matlab program, the upper-right side of the image was removed through a sorting and cutting procedures acting on the raw data. At this stage we had to guarantee that all the features computed were independent from noise. Then, by using the Image Processing toolbox, the centroid, the principal inertia axis and then the feature set (L,W,α) were calculated for each record. To take a look at the separation of the 12 classes provided by the proposed feature vectors, in Fig. 3 a tridimensional scatter graph is reported. There is a good separation among classes. This indicates that the feature vectors obtained are able to yield good classification from the classifiers.

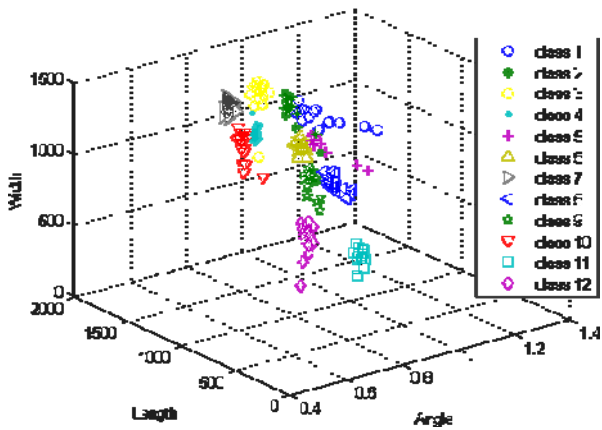

Fig. 3. Tridimensional scatter graph of the proposed feature vectors for 12 classes.

*C. Classifiers*

To evaluate the proposed shape descriptor system as recognition of structure defect we tested it with three different machine-learning based classifiers: J48 decision trees, Multilayer Perceptron neural networks and Naive Bayes [36].

The classification performance of each classifier is evaluated using the ten-fold cross-validation method [37], a model validation technique for assessing how the classification results will generalize for an independent data set. The 12 different classes are derived from the overall dataset by comparing each defect to the others. Then, all the data have been randomly divided into 10 disjoint subsets (folders), each containing approximately the same amount of instances. In each experiment, nine folders have been used as training data, while the remaining folder is used as validation. This process has been repeated 10 times, for each different choice of the validation folder.

The performance results, reported in Table I as mean values, are presented in terms of accuracy, sensitivity, specificity, precision and Matthews correlation coefficient [38].

TABLE I. EXPERIMENTAL RESULTS

| Classifier | Mean values of the "ten-fold cross validation" results | | | | |
|---|---|---|---|---|---|
| | *Accuracy* | *Sensitivity* | *Specificity* | *Precision* | *Matthews* |
| J48 | 0.96 | 0.74 | 0.98 | 0.81 | 0.75 |
| Naive Bayes | 0.95 | 0.68 | 0.97 | 0.67 | 0.63 |
| Multilayer Perceptron | 0.98 | 0.85 | 0.99 | 0.89 | 0.84 |

The experiment results confirm the good class separation shown in Fig. 3. In particular, the performance parameters related to the neural network were very high, near to 1.

V. CONCLUSIONS

In this paper we have investigated the application of a "CBIR" method to characterize aerospace structure defects based on eddy current testing. We have used the shape that assumes the impedance of the EC probe coil in the plane complex as signatures to identify changes in the test piece. We have made use of information related to the boundary of the image through a feature vector composed by only three geometric parameters: length, width and orientation angle of the shape. To evaluate the suitability and usefulness of the proposed set of features we tested them with three different machine-learning based classifiers: J48 decision trees, Multilayer Perceptron neural network and Naive Bayes. The results, evidencing an accuracy rate of almost 100%, demonstrates the effectiveness of our approach.

ACKNOWLEDGMENTS

This work has been supported in part by Distretto Aerospaziale della Campania (DAC) in the framework of the CERVIA project - PON03PE_00124_1.